\documentclass{article}
\usepackage{arxiv}
\let\undertitle\relax  % Disable the undefined command
\newcommand{\undertitle}{} % Leave empty or add subtitle
\newcommand{\headeright}{TRACES Submission}
\pdfoutput=1
\usepackage{textgreek} 
\usepackage[utf8]{inputenc} % allow utf-8 input
\usepackage[T1]{fontenc}    % use 8-bit T1 fonts
\usepackage{hyperref}       % hyperlinks
\usepackage{url}            % simple URL typesetting
\usepackage{booktabs}       % professional-quality tables
\usepackage{amsfonts}       % blackboard math symbols
\usepackage{nicefrac}       % compact symbols for 1/2, etc.
\usepackage{microtype}      % microtypography
\usepackage{lipsum}		% Can be removed after putting your text content
\usepackage{graphicx}
\usepackage{natbib}
\usepackage{doi}
\usepackage{amsmath}
\usepackage{color}
\usepackage{caption}
\usepackage{pifont}
\newcommand{\cmark}{\ding{51}}%

\title{\textbf{TRACES: Temporal Recall with Contextual Embeddings for Real-Time Video Anomaly Detection}}

\author{
  \textbf{Yousuf Ahmed Siddiqui}\thanks{Corresponding author. Email: \texttt{K214594@nu.edu.pk}} \\
  \textbf{Sufiyaan Usmani} \\
  \textbf{Umer Tariq} \\
  Department of Computer Science \\
  FAST-NUCES \\
  Karachi, Pakistan \\
  \texttt{\{K214594, K213195, K213432\}@nu.edu.pk}
  \AND
  \textbf{Dr. Jawwad Ahmed Shamsi} \\
  \textbf{Dr. Muhammad Burhan Khan} \\
  System Research Laboratory (SysLab) \\
  FAST-NUCES \\
  Karachi, Pakistan \\
  \texttt{\{jawwad.shamsi, burhan.khan\}@nu.edu.pk}
}

\makeatletter
\@ifundefined{bibtex}{}{\immediate\write18{bibtex template}}
\makeatother

\renewcommand{\shorttitle}{TRACES: Temporal Recall with Contextual Embeddings}

\hypersetup{
pdftitle={TRACES: Temporal Recall with Contextual Embeddings for Real-Time Video Anomaly Detection},
pdfauthor={Yousuf Ahmed Siddiqui, Sufiyaan Usmani, Umer Tariq, Jawwad Ahmed Shamsi, Muhammad Burhan Khan},
pdfkeywords={Video anomaly detection, Real-time processing, Contextual embeddings, Weakly supervised learning, Temporal modeling}
}
\begin{document}
\maketitle

\begin{abstract}
Video anomalies often depend on contextual information available and temporal evolution. Non-anomalous action in one context can be anomalous in some other context. Most anomaly detectors, however, do not notice this type of context, which seriously limits their capability to generalize to new, real-life situations. Our work addresses the context-aware zero-shot anomaly detection challenge, in which systems need to learn adaptively to detect new events by correlating temporal and appearance features with textual traces of memory in real time. Our approach defines a memory-augmented pipeline, correlating temporal signals with visual embeddings using cross-attention, and real-time zero-shot anomaly classification by contextual similarity scoring. We achieve 90.4\% AUC on UCF-Crime and 83.67\% AP on XD-Violence, a new state-of-the-art among zero-shot models. Our model achieves real-time inference with high precision and explainability for deployment. We show that, by fusing cross-attention temporal fusion and contextual memory, we achieve high fidelity anomaly detection, a step towards the applicability of zero-shot models in real-world surveillance and infrastructure monitoring.
\end{abstract}

% keywords can be removed
\section{Introduction}
Detecting anomalies in video without any previous exposure to anomalous instances is a fundamental problem for surveillance, industrial monitoring, and safety systems \cite{zhu2024advancingvideoanomalydetection}. The majority of current zero-shot anomaly detection (ZSAD) algorithms utilize vision-language models such as CLIP, pseudo-anomaly awareness, prompt learning, or multi-scale feature aggregation in order to generalize to unknown anomaly types \cite{AA-CLIP2025, AF-CLIP2025, PA-CLIP2025, KAnoCLIP2025, AdaCLIP2024, AnomalyCLIP2024, DualImageCLIP2024, khan2025contextawarezeroshotanomalydetection}. Neuroscience provides a teaching analogy: learning episodes create long-lasting traces in the brain, e.g., the "motor cortex trace" found in monkeys when acquiring novel behaviors, that persist even when performing routine actions \cite{Losey2024MemoryTrace}. Guided by this maxim, we suggest modeling anomalous and non-anomalous past contexts as "traces" held in a memory bank outside the brain, to which conditional access is given based on the current scene. We present a new zero-shot video anomaly detection approach that combines temporal recall and contextual embeddings, employing cross-attention to combine motion and appearance features into the CLIP embedding space, keeping memory banks of both anomalous and non-anomalous contexts, and conducting anomaly scoring through similarity comparison with textual context vectors. Experimental performance reveals our method outperforms current zero-shot baselines on typical metrics such as AUC-ROC and F1-score at low latency \cite{AA-CLIP2025, gao2025adaptclipadaptingclipuniversal, khan2025contextawarezeroshotanomalydetection}. Ablation experiments investigate memory bank capacity, temporal windowing, and varying fusion mechanisms and affirm that adding contextually relevant traces has dramatic detection performance improvement under real-world open-vocabulary and unseen anomaly type constraints \cite{khan2025contextawarezeroshotanomalydetection, AF-CLIP2025}.

Despite breakthroughs like zero-shot VAD models (Flashback \cite{flashback2025}) and memory-augmented appearance-motion networks (AMSRC \cite{amsrc2022}, PDMNet \cite{pdmnet2024}) have been made, no single framework as yet exists that consistently recalls contextually appropriate anomalous "traces" in different scenes without the need for previous exposure to all types of anomalies. Current approaches commonly suffer from one or more of the following limitations: they commonly have a lack of temporal modeling that resolves long-range dependencies (and therefore fail on subtle or slowly changing anomalies) for instance, weakly-supervised approaches like RTFM \cite{rtfm2021} handle some temporal dependencies but still require domain-specific exposure, they fail to combine appearance and temporal semantics in a manner that maintains contextual relevance across environments, resulting in false positives/negatives when scene context changes (as exemplified in appearance-motion consistency models like AMSRC \cite{amsrc2022}). Most methods are highly reliant on labels (weakly supervised or supervised) or need tuning of regular patterns per deployment environment, which restricts generalizability and practical use. Zero-shot methods such as Flashback \cite{flashback2025} decrease this reliance but never explicitly combine appearance memories with learned prototypes to respond to environment changes. We introduce TRACE (Temporal Recall with Contextual Embeddings), a new zero-shot video anomaly detection system that combines memory, motion, appearance, and contextuality to overcome the shortcomings of existing methods \cite{rtfm2021,flashback2025}. TRACE consists of four major components:

\begin{itemize}
    \item {Context-Memory Bank}, which stores anomalous and non-anomalous trace embeddings, allowing retrieval of contextual priors.
    \item {Motion–Appearance Fusion Module}, utilizing temporal cross-attention \cite{mgfn2021} to couple dynamic behavioral patterns with visual semantics.
    \item {Zero-Shot Anomaly Scoring Mechanism}, which predicts anomaly likelihood through similarity between fused embeddings and textual context vectors without using anomaly-labeled data during training \cite{cliptsa2023}.
    \item {Optimized inference pipeline}, for real-time deployment.
\end{itemize}

Motivated by cognitive retrieval mechanisms—demonstrated by Flashback's memory-guided recall before response \cite{flashback2025}—TRACE generalizes this framework by making recall dependent on present contextual clues and simultaneously modeling motion in addition to appearance, thus improving accuracy, contextual stability, and zero-shot transfer while being computationally efficient.

The rest of this paper is structured as follows. Section II summarizes prior work on zero-shot anomaly detection \cite{cliptsa2023,flashback2025} and context-aware video understanding \cite{mgfn2021,rtfm2021}, pointing out the weaknesses of supervised and fusion-based models. Section III presents the TRACE framework that is being proposed, explaining the contextual memory bank, motion–appearance fusion through temporal cross-attention \cite{mgfn2021}, and zero-shot anomaly scoring mechanism \cite{cliptsa2023}. Section IV presents the experimental configuration, such as datasets (UCF-Crime, XD-Violence), metrics for evaluation, and implementation details, along with baseline approaches applied to compare with. Lastly, Section VI concludes with a discussion on contributions, limitations, and directions for future work towards real-world deployment of context-aware zero-shot anomaly detection.

\section{Related Work}
\subsection{Fully-Supervised and Weakly-Supervised Methods}
Fully supervised VAD assumes all frames are annotated as normal or anomalous. This is rarely practical since anomalies are inherently scarce. Consequently, fully supervised VAD often boils down to regular video classification, as also observed in violence detection applications \cite{Wu2024}. Rather, most research considers VAD as either semi-supervised (normal training data only) or weakly-supervised (video-level annotation only). Semi-supervised VAD trains on normal videos only. Early deep approaches in this paradigm employ reconstruction or forecasting: e.g., convolutional autoencoders (ConvAE) or future-frame predictors learn a low-dimensional normality model such that anomalies have large reconstruction/prediction errors \cite{Wu2024}. More recent methods augment these models with complex pretext tasks. Huang et al. \cite{Huang2022} introduce a two-stream encoder that imposes semantic consistency on appearance and motion representations of normal frames, thereby making anomalies (with semantically inconsistent appearance-motion features) prominent. Lu et al. \cite{Lu2024} present PDM-Net, retaining prototypical dynamic normal event patterns during inference, brief video segments are compared to learned normal-motion prototypes in memory to predict frames, facilitating abnormal motions to be detected more easily. Overall, semi-supervised VAD approaches depend on learning a "normal pattern" through self-supervised tasks (e.g. reconstruction, frame prediction, or contrastive learning) and marking away from this norm \cite{Wu2024,Huang2022,Lu2024}.
Weakly-supervised VAD only gets coarse labels (normal or abnormal) at the video level, without frame-level annotations. One prevalent paradigm is multiple-instance learning (MIL) on video snippets. The model learns to give high anomaly scores to certain snippets within videos that are labeled anomalous. For instance, Tian et al.'s RTFM \cite{Tian2021} creates a new MIL loss with focus on feature magnitude to enhance subtle anomalies, with big gains on benchmarks. Zhong et al. \cite{Zhong2021} employ a multi-scale graph convolutional network that combines snippet features over time, enhancing temporal localization under weak supervision. These and related works continually enhance snippet-level detection by capturing temporal context (e.g. through attention or graph modules) under the MIL paradigm. Interestingly, more recent weakly-supervised approaches have come to include large pre-trained encoders e.g., Joo et al. \cite{Joo2023} make use of CLIP's vision transformer representations with a learned Temporal Self-Attention (CLIP-TSA), improving discriminability, Semi- and weakly-supervised VAD approaches make use of either solely normal data or video-level annotations to learn normality. They usually concentrate on reconstruction/prediction networks, self-supervised objectives, and MIL-based ranking, but all need some domain-specific training data \cite{Wu2024,Tian2021,Zhong2021}.

\subsection{Unsupervised Open-set and Zero-shot Methods}
In unsupervised VAD, no labels are ever employed in training and the model might even get no regular videos anomaly detection is based solely on intrinsic signals. Conventional unsupervised methods train on regular data (or do not use any data) and identify anomalies as statistical outliers. For instance, one-class models and generative networks (such as autoencoders, generative flows) are learned on typical video and signal high reconstruction error as anomalies \cite{Wu2024,Huang2022}. Methods like appearance-motion consistency \cite{Huang2022} and prototypical memory banks \cite{Lu2024} belong to this category where they learn to represent normal patterns such that deviations (in consistency or prototype matching) signal abnormal events. These unsupervised models often extend to open-set VAD, where a few seen anomaly types are available during training: in open-set VAD the goal is to detect unseen anomalies beyond the labeled classes \cite{Wu2024}. Open-set methods typically use specialized losses or margin learning to separate normal, seen-anomalous, and unknown-anomalous distributions, but this area is still emerging \cite{Wu2024}.
A newer frontier is zero-shot VAD using large vision–language models. These approaches have no target-domain training data. They use models such as CLIP or vision-language models to associate video clips with semantic descriptions. For example, "caption-and-score" pipelines (such as LAVAD) caption each clip of a video first using a visual-language model and then pass text through a large language model to score anomalousness. While effective, autoregressive captioning is slow. More recent contributions directly adapt CLIP. Several strategies have been suggested: Ma et al.'s AA-CLIP \cite{Ma2025} injects anomaly-aware prompts into CLIP; Fang et al.'s AF-CLIP \cite{Fang2025} learns prompt embeddings anomaly-centered; Pan et al.'s PA-CLIP \cite{Pan2025} suggests pseudo-anomaly guidance; and Li et al.'s KanoCLIP \cite{Li2025} involves knowledge-driven prompt learning as well as cross-modal fusion. There are other variants such as hybrid prompt tuning (Ada-CLIP \cite{Cao2024}), object-agnostic prompts (AnomalyCLIP \cite{Zhou2025}), dual-image ensembles \cite{Zhang2024}, and spatio-temporal contrastive learning (Khan et al. \cite{Khan2025}). Gao et al. \cite{Gao2025} also demonstrate that fine-tuning or learning prompts on CLIP results in a more "universal" anomaly detector. All these zero-shot approaches based on CLIP have competitive accuracy and even generate textual explanations, albeit at the cost of dataset-agnostic training.

\section{Methodology}
\begin{figure}[t]
  \centering
  \includegraphics[width=0.55\columnwidth]{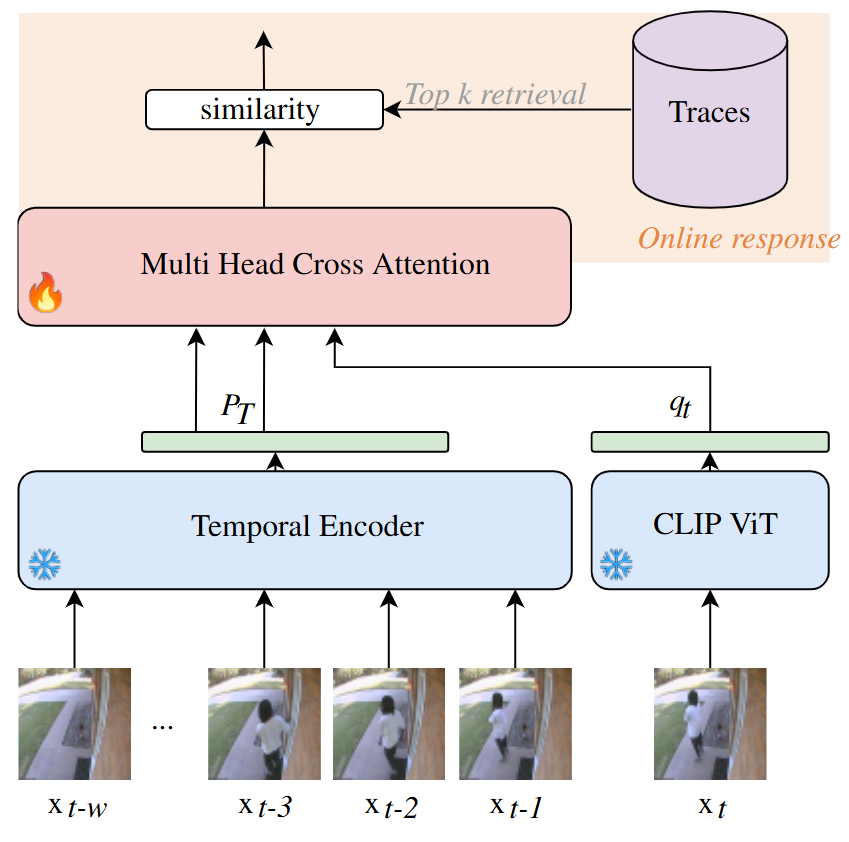}
  \caption{Proposed framework for anomaly detection.}
  \label{fig:framework}
\end{figure}

\subsection{Architecture overview}
The suggested TRACE architecture combines frozen pretrained encoders with light-weight adapter modules to facilitate effective temporal contextual reasoning as shown in Figure~\ref{fig:framework}, which shows a high-level view of the model's components and information flow.

The CLIP vision encoder $E_{\text{vis}}$ extracts frame-level appearance embeddings $f_t \in \mathbb{R}^d$ from the current frame $x_t$ \cite{radford2021clip}. Simultaneously, a frozen temporal encoder $E_{\text{temp}}$, implemented as TimeSformer \cite{bertasius2021timesformer}, processes a short sequence of preceding frames $\{x_{t-W}, \ldots, x_{t-1}\}$ to capture temporal dynamics, yielding temporal representations $r_t \in \mathbb{R}^{k}$.

To facilitate cross-modal alignment between the visual and temporal representations, we propose lightweight adapter modules, denoted as \textit{Atemp}, guided by parameter-efficient tuning principles \cite{houlsby2019adapter,hu2022lora}. The adapters map the frozen embeddings into a common latent subspace:
\begin{equation}
    q_t = A_{\text{vis}}(f_t), \quad P_T = A_{\text{temp}}(r_t),
\end{equation}
where $q_t, P_T \in \mathbb{R}^{d'}$, and $d'$ is often smaller than $d$ to cut down computation overhead. Both adapters apply Layer Normalization and dropout regularization to ensure stability.

\begin{figure}[t]
  \centering
  % ---- Left Figure ----
  \begin{minipage}[t]{0.48\columnwidth}
    \centering
    \includegraphics[width=\textwidth]{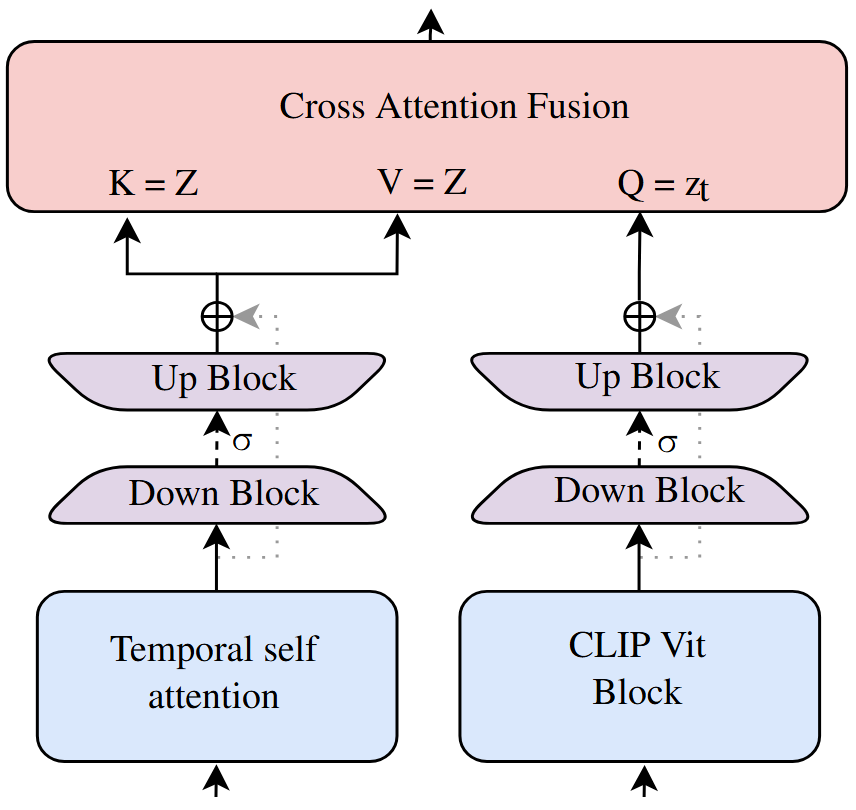}
    \caption{Architecture integrates CLIP-ViT for visual-language representation, with up/down adapter blocks and Temporal self-attention to capture sequence dynamics, while cross-attention fusion aligns multi-modal features for contextual anomaly reasoning.}
    \label{fig:illus_2}
  \end{minipage}%
  \hfill
  % ---- Right Figure ----
  \begin{minipage}[t]{0.48\columnwidth}
    \centering
    \includegraphics[width=\textwidth]{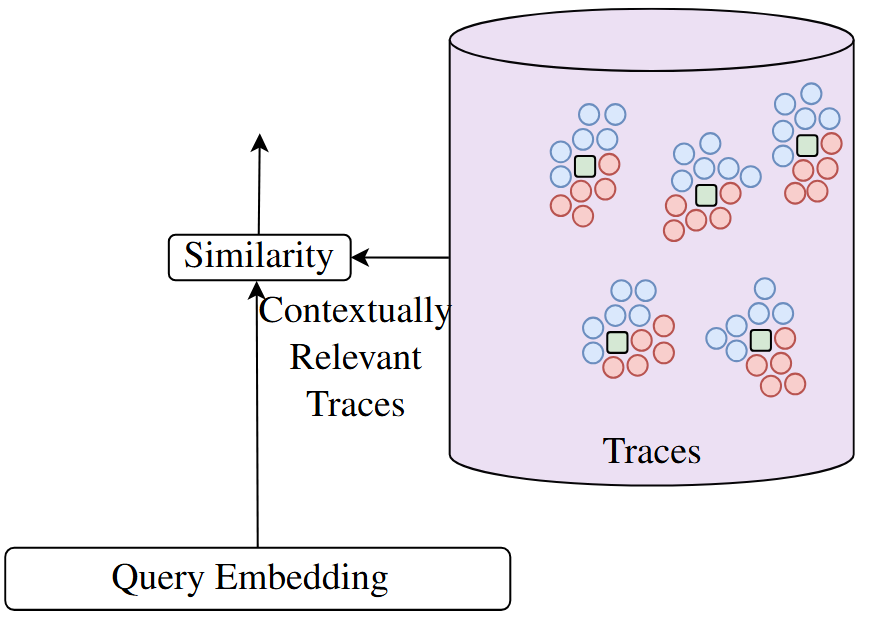}
    \caption{Simplified scheme of the proposed trace retrieval framework, showing how a query embedding is compared against context-specific trace vectors in the Trace Bank.}
    \label{fig:illus_3}
  \end{minipage}
\end{figure}

The fusion mechanism is achieved through multi-head temporal cross-attention module \cite{vaswani2017attention}. In this, queries are obtained from the appearance embedding $Q_T$, while the keys and values are taken from the temporal features $P_T$, The structure of the proposed fusion mechanism is illustrated in Figure~\ref{fig:illus_2}. This design enables TRACE to adaptively combine frame-level semantics with temporal consistency cues, resulting in a fused representation $U_T \in \mathbb{R}^{d'}$ that encodes both contextual and appearance information:
\begin{equation}
    U_T = \text{CrossAttn}(q_t, P_T, P_T).
\end{equation}

This attention is applied within a sliding temporal window of size $W$ (e.g., $W=16$ or $32$). An enlarged window captures long-range dependencies, but adds latency, while a smaller one favors responsiveness. 

All heavy pretrained backbones (CLIP encoders and motion encoder) are frozen. 
The only trainable elements are the light-weight adapters $A_{\text{vis}}, A_{\text{temp}}$ and the 
cross-attention block. This structure is parameter-efficient 
and maintains generalization to novel anomaly types as prescribed by the zero-shot learning philosophy \cite{xian2018zeroshot,radford2021clip}.

\subsection{Traces Bank}
Traces are textual representations of contextual environments (e.g., \textit{school corridor, kitchen, hospital, parking lot}) that capture both anomalous and non-anomalous scene stories. Each trace is a contextually relevant anomalous or non-anomalous vector outlining a real-world scenario or event, mapped into a joint vision–language space using the frozen CLIP text encoder $E_{\text{text}}$ \cite{radford2021clip}.

To encode varied semantic conditions, the total number of 70 unique contextual settings was taken into account, and each of these had more than one anomalous and non-anomalous trace. A total of about one million anomalous and non-anomalous events were created and embedded with the CLIP text encoder and collectively have an embedding size of about 2.05 GB (based on 1M embeddings $\times$ 512 dimensions $\times$ 4 bytes per float).

% idhr tak hwa he thursday
Traces were generated with a two-stage pipeline. First, the LLaMA 4.1 (128-expert) model on GroqCloud, was queried to create a range of different types of scenes and location categories. In the second phase, for each setting recognized, the same model was queried to create 5--7 anomalous and  {5--7 non-anomalous} text descriptions that define realistic activities or events that can occur in that setting. This method enabled the creation of highly contextualized and semantically harmonious text scenes, establishing a pseudo-linguistic memory of actual behavioral patterns.

Each trace $t_i$ is embedded into the CLIP space as $E_{\text{text}}(t_i)$ and stored in a high-capacity memory bank for retrieval. For scalability and efficiency, we use a FAISS-based vector database \cite{johnson2019faiss} with clustering and redundancy suppression:
\begin{itemize}
    \item Highly similar traces are merged into centroids to preserve representativeness.
    \item Redundant embeddings are pruned to maintain semantic diversity across contexts.
\end{itemize}

At inference time, the query embedding $u_t$ (derived from the combined temporal-appearance feature through cross-attention) is matched against the full set of context vectors. The top-$k$ most similar traces are retrieved separately for anomalous ($T_A$) and non-anomalous ($T_N$) subsets via cosine similarity:
\begin{equation}
    \text{Recall}(u_t) = \operatorname{TopK}\big( \cos(u_t, \, T_A \cup T_N) \big).
\end{equation}
This context-aware retrieval mechanism allows the model to reason across semantically comparable scenarios and not raw feature distances, enhancing discrimination in challenging scenarios. That is, traces serve as pseudo scene memory, informing the system to contextualize the current embedding in relation to contextually comparable instances.

\begin{figure}[t]
  \centering
  \vspace{0pt} % ensures top alignment
  \begin{minipage}[t]{0.63\columnwidth}
    \vspace{0pt} % normalize top alignment
    \subsection{Zero-Shot Anomaly Scoring and Inference Pipeline}
    Given a fused spatio-temporal embedding $u_t$, the goal is to make anomaly likelihood inference without being exposed to any anomalous training samples. To do this, we find the semantic alignment of $u_t$ with anomalous and non-anomalous trace embeddings in the contextual memory bank. We use cosine similarity as the similarity metric due to its scale invariance and effectiveness in heterogeneous representation alignment as well as retrieval-based inference~\cite{jegou2011product,schroff2015facenet}.
    
    For the anomalous trace subset:
    \begin{equation}
        s_A = \max_{i \in \text{Top-}k(T_A)} \cos(u_t, T_{A,i}),
    \end{equation}
    and similarity for non-anomalous traces:
    \begin{equation}
        s_N = \max_{j \in \text{Top-}k(T_N)} \cos(u_t, T_{N,j}),
    \end{equation}
    
  \end{minipage}
  \hfill
  \begin{minipage}[t]{0.33\columnwidth}
    \vspace{0pt} % aligns top of this box to top of the left box
    \centering
    \includegraphics[width=\linewidth]{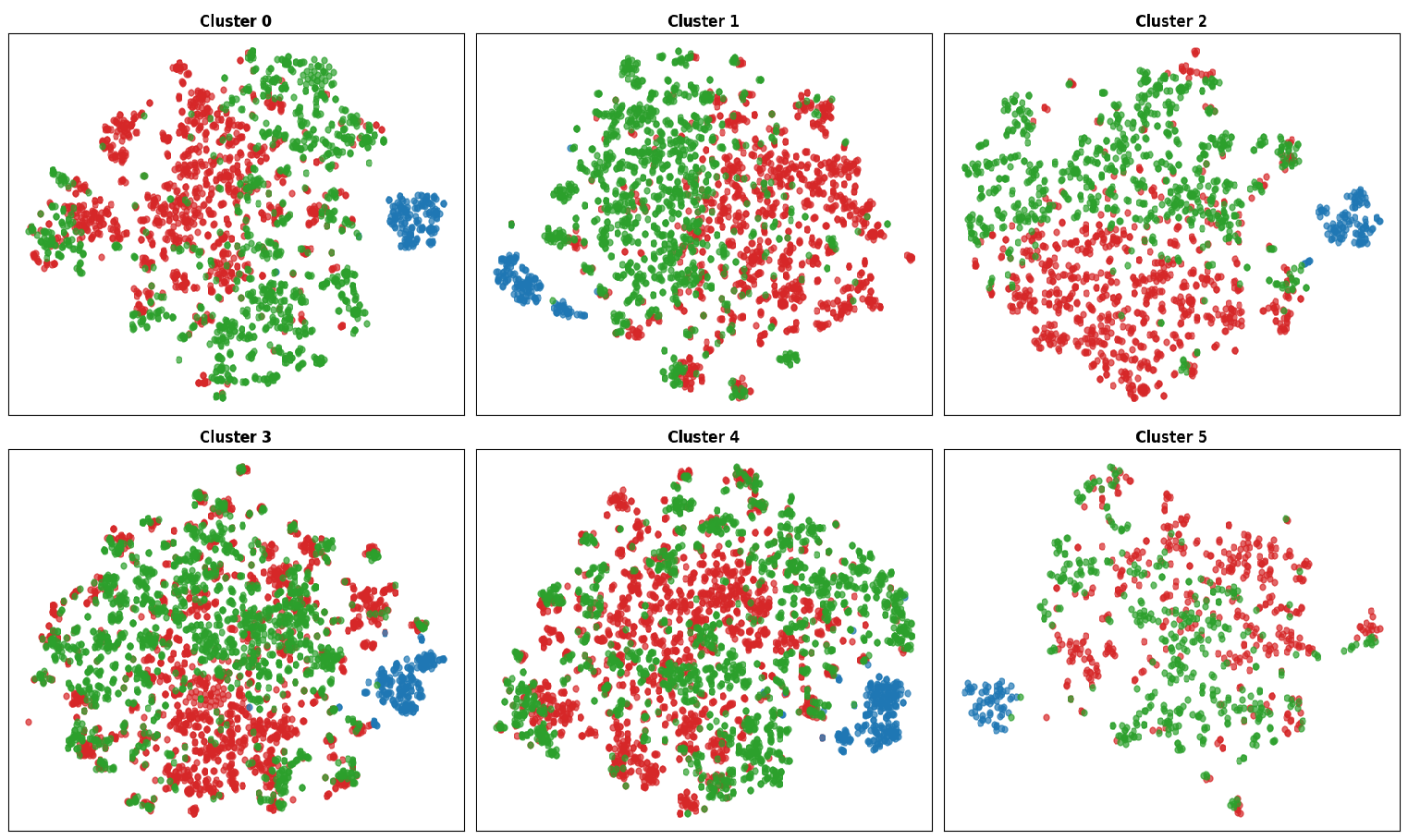}
    \caption{t-SNE visualization of clustered trace embeddings from the Traces Bank. Six distinct context clusters are observed, each exhibiting different distributions of anomalous (red) and non-anomalous (blue) vectors.}
    \label{fig:embeddings}
  \end{minipage}
\end{figure}

Figure~\ref{fig:illus_3} gives an overview of the  {trace generation and retrieval pipeline}, illustrating how textual descriptions are converted into CLIP embeddings and utilized for runtime matching.  
Furthermore, Figure~\ref{fig:embeddings} plots the  {t-SNE clustering} of the trace bank, where each cluster has a heterogeneous distribution of anomalous and non-anomalous traces. The clusters illustrate that the embeddings separate naturally by semantic context and keep anomaly-aware local structure intact.

where $\cos(\cdot, \cdot)$ is cosine similarity in the joint embedding space. Other aggregation approaches, e.g., softmax-weighted similarity or attention pooling, can also be used to reduce noise in nearest-neighbour retrieval~\cite{liu2018ano_pred,tian2021rtfm}.
    
    This retrieval-guided reasoning enables the framework to infer anomaly likelihood in a zero-shot manner by extracting context-conditioned similarity patterns instead of explicit supervision.

\subsection{Score Aggregation and Calibration}
The anomaly score $S_t$ is defined as a discriminative difference between anomalous and non-anomalous similarities:
\begin{equation}
    S_t = s_A - s_N,
\end{equation}
or alternatively as an aggregate additive measure:
\begin{equation}
    S_t = s_A + s_N + \epsilon,
\end{equation}
where $\epsilon$ is a bias term for calibration.  
A softmax-normalized version can also produce probabilistic confidence scores \cite{pang2021deep}. The model is chosen for empirical stability and interpretability on large-scale benchmarks.

A binary classification decision threshold $\\theta$ is used next:
\begin{equation}
    \text{Label}(t) =
    \begin{cases}
        \text{Anomalous}, & S_t \geq \theta, \\
        \text{Normal}, & S_t < \theta.
    \end{cases}
\end{equation}
The threshold $\theta$ can optionally be globally set or tuned on a minimal validation set, as in previous weakly- and zero-shot anomaly detection researches \cite{tian2021rtfm,zhong2023lavad}.

For efficiency, precomputed trace embeddings are indexed and searched with FAISS-based vector search \cite{johnson2019faiss}, and appearance–temporal fusion is performed online. The average per-frame latency is shown in implementation details to provide reproducibility.

Because CLIP and temporal features can vary in distribution and magnitude before adapter projection, dropout and Layer Normalization are used to stabilize optimization and preserve cross-modal alignment \cite{vaswani2017attention}. In addition, cosine similarities are temperature-scaled to enhance score separability:

\begin{equation}
    \cos_\tau(u, v) = \frac{\cos(u, v)}{\tau},
\end{equation}
where $\tau$ is a temperature hyperparameter that enhances the calibration margin between anomalous and non-anomalous responses, thus enhancing robustness in open-set conditions \cite{guo2017calibration}.

\newpage

\section{Experiments}
\subsection{Experimental Setup}
We test TRACE on two popular video anomaly detection (VAD) benchmarks:  {UCF-Crime} \cite{sultani2018real} and  {XD-Violence} \cite{wu2020not}.  
\begin{itemize}
    \item  {UCF-Crime:} A massive untrimmed surveillance video dataset of $13$ anomaly classes like robbery, accident, and abuse. Annotations are given at the frame level. Frames were sampled at $30$ fps and resized to a constant resolution as done before in work \cite{tian2021weakly, zhong2019graph}.
    \item  {XD-Violence:} Includes long untrimmed videos of violent and non-violent activity, annotated at the segment level. As with standard procedure \cite{wu2020not}, we preprocessed by normalizing all the clips to $30$ fps.
\end{itemize}

\noindent
\begin{minipage}[t]{0.58\textwidth}
\raggedright
Assessment on both datasets allows us to examine TRACE's generalizability across a range of anomaly classes (crime, accidents, violence). Appearance embeddings were obtained from the frozen CLIP visual encoder $E_{vis}$~\cite{radford2021learning}, and temporal embeddings were extracted from a pretrained temporal backbone $E_{temp}$ (TimeSformer~\cite{bertasius2021timesformer}, frozen). Temporal features were extracted over short clips to get local temporal dynamics. All embeddings were $\ell_2$-normalized before adapter projection to achieve scale-invariant similarity comparisons.  

Adapter modules $A_{vis}$ and $A_{mot}$ were realized as two-layer MLPs projecting into a shared $512$-dimensional latent space. Feature fusion was achieved through a single-layer temporal cross-attention module with $8$ heads (head dimension $64$), enabling modality alignment and temporal context aggregation. During inference, the top-$5$ nearest neighbors were obtained with cosine similarity. This retrieval configuration adheres to memory-augmented paradigms applied in anomaly detection~\cite{park2020learning, doshi2022memorizing}.  

We present AUC-ROC and F1-score at both frame- and segment-level, as per previous VAD literature~\cite{sultani2018, tian2021weakly, wu2020not}. A frame $t$ was labeled anomalous if its anomaly score $S_t \geq \theta$.
\end{minipage}%
\hfill
\begin{minipage}[t]{0.38\textwidth}
\centering
\vspace{0pt} % aligns top edges perfectly
\includegraphics[width=\linewidth]{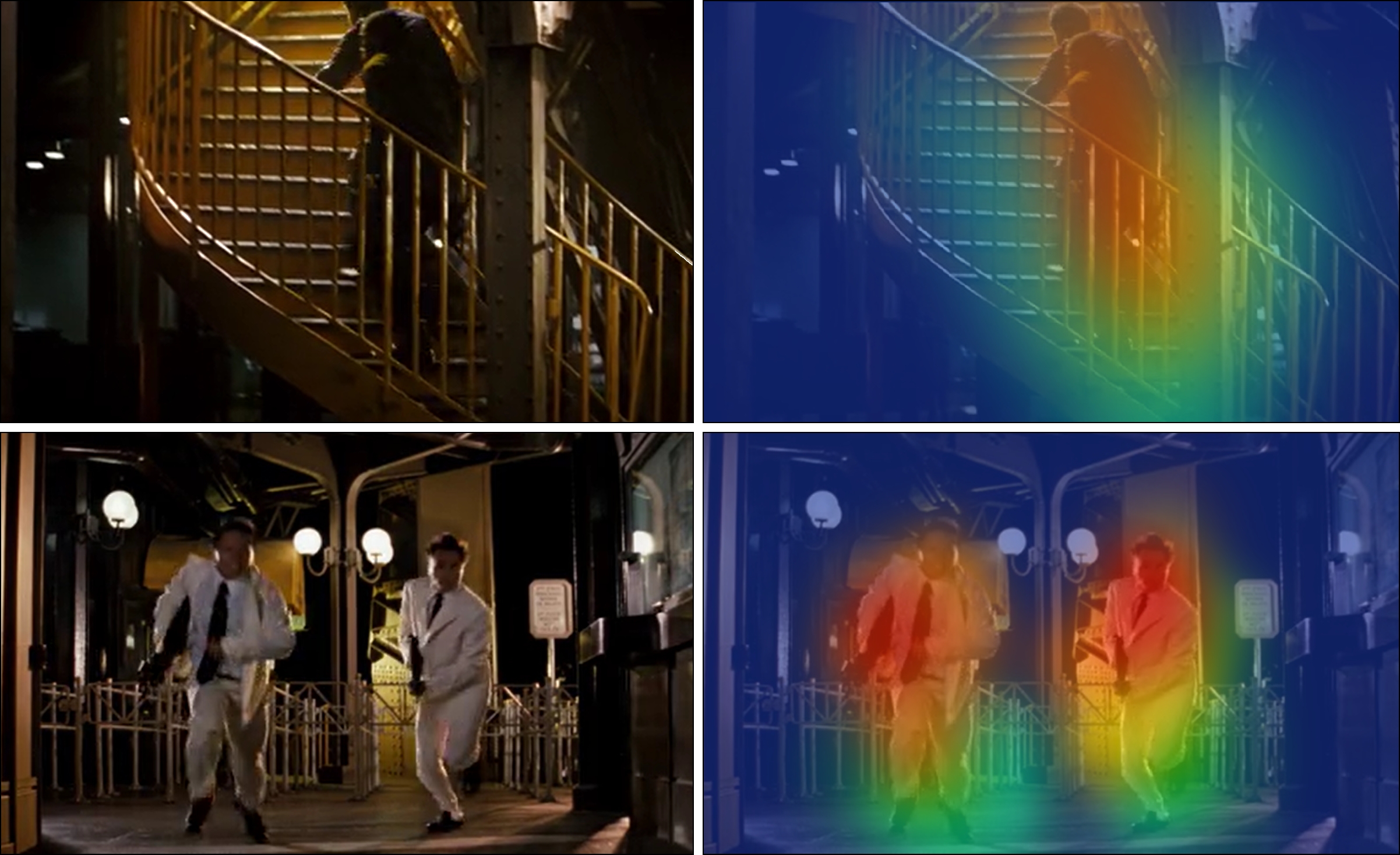}
\captionof{figure}{
Qualitative visualization of cross-attention interpretability on the XD-Violence dataset. 
The top frame shows a non-anomalous instance, while the bottom frame shows an anomalous event.
Grad-CAM-inspired cross-attention heatmaps emphasize the spatial and temporal regions most influential to the model’s zero-shot reasoning.
}
\label{fig:attention}
\end{minipage}

\subsection{Comparision with SOTA methods}
We compare TRACE to representative baselines across zero-shot, weakly-supervised, and unsupervised paradigms. These consist of recent CLIP-based variants and traditional reconstruction-driven methods \cite{flashback} \cite{aa-clip} \cite{sultani2018, tian2021weakly} \cite{hasan2016learning, gong2019memorizing}

This choice represents the primary methodological dimensions in VAD: prompt-tuned zero-shot prediction, weakly-labeled discriminative training, and unsupervised reconstruction-based methods.  

In contrast, baselines tend to fail in adverse scenarios like low lightning, occlusion, or fast scene transition \cite{wu2020not, zhong2019graph}. Robustness in TRACE is preserved via retrieval-augmented inference and contextualized memory alignment that allows for fine-grained anomaly attribution.

Qualitative examples from XD-Violence are shown in Figure~\ref{fig:attention}. The topmost frame is a non-anomalous sample and the lower frame is an anomalous sample that we also show Grad-Cam styled \cite{camgrad} cross-attention heat maps for interpretibility.

\begin{table}[t]
\caption{Comparison against state-of-the-art video anomaly detectors on UCF-Crime and XDViolence. Methods are divided into supervision level (weakly-supervised, one-class, unsupervised, and zero-shot). 
TRACE has the highest accuracy on both datasets and is the first method that is concurrently zero-shot, real-time, and explainable. 
Bold numbers indicate the best result.}
\centering
\footnotesize
\begin{tabular}{lccccc}
\hline
 {Method} &  {Expl} &  {RT} &  {UCF AUC} &  {XD AP} &  {XD AUC} \\
\hline
\multicolumn{6}{c}{ {Weakly-Supervised}} \\
\hline
Sultani et al. \cite{sultani2018} & - & \cmark & 77.92 & - & - \\
GCL \cite{GCL} & - & \cmark & 79.84 & - & - \\
Wu et al. \cite{Wu2020} & - & \cmark & 82.44 & 73.20 & - \\
RTFM \cite{RTFM} & - & \cmark & 84.03 & 77.81 & - \\
Wu and Liu \cite{WuLiu2020} & - & \cmark & 84.89 & 75.90 & - \\
MSL \cite{MSL} & - & \cmark & 85.62 & 78.58 & - \\
S3R \cite{S3R} & - & \cmark & 85.99 & 80.26 & - \\
MGFN \cite{MGFN} & - & \cmark & 86.98 & 80.11 & - \\
CLIP-TSA \cite{CLIPTSA} & \cmark & - & 87.58 & 82.17* & - \\
VadCLIP \cite{VadCLIP} & \cmark & - & 88.02 & 84.51 & - \\
Holmes-VAD \cite{HolmesVAD} & \cmark & - & 84.61$^\dagger$ & 84.96$^\dagger$ & - \\
VERA \cite{VERA} & \cmark & - & 86.55 & 70.54 & 88.26 \\
\hline
\multicolumn{6}{c}{ {One-Class}} \\
\hline
Hasan et al. \cite{Hasan2016} & - & \cmark & - & - & 50.32 \\
Lu et al. \cite{Lu2019} & - & \cmark & - & - & 53.56 \\
BODS \cite{BODS} & - & \cmark & 68.26 & - & 57.32 \\
GODS \cite{BODS} & - & \cmark & 70.46 & - & 61.56 \\
\hline
\multicolumn{6}{c}{ {Unsupervised}} \\
\hline
GCL \cite{GCL} & - & \cmark & 74.20 & - & - \\
Tur et al. \cite{Tur2019a} & - & \cmark & 65.22 & - & - \\
Tur et al. \cite{Tur2019b} & - & \cmark & 66.85 & - & - \\
DyAnNet \cite{DyAnNet} & - & \cmark & 79.76 & - & - \\
RareAnom \cite{RareAnom} & - & \cmark & - & - & 68.33 \\
\hline
\multicolumn{6}{c}{ {Zero-Shot}} \\
\hline
LAVAD \cite{LAVAD} & \cmark & - & 80.28 & 62.01 & 85.36 \\
Flashback-PE \cite{fb2025} & \cmark & \cmark & 87.29 & 75.13 & 90.54 \\
 {TRACE (Ours)} & \cmark & \cmark &  \textbf{90.40} &  \textbf{83.67} &  \textbf{92.15} \\
\hline
\end{tabular}
\label{tab:sota_comparison}
\end{table}

\subsection{Ablation Analysis}

For systematically measuring the contribution of every component of  {TRACE} in terms of its ablation, we thoroughly perform an ablation study on both the UCF-Crime and XD-Violence datasets. Our analysis revolves around three primary architectural features:
(i) modality-specific adapter projections,  
(ii) temporal cross-attention fusion, and  
(iii) the contextual trace memory bank.  
Also, we test the hyper-sensitivity to memory bank cardinality, top-$k$ retrieval size, and temporal receptive field (window length).

\paragraph{Effect of Temporal Cross-Attention.}  
Substitution of cross-attention in the proposed architecture with naive concatenation fusion (Concat-Fusion) leads to a drastic loss in performance, affirming that structured temporal alignment between appearance and motion streams is vital. Cross-attention selectively suppresses background noise while highlighting salient temporal relationships.

\begin{table}[h]
\centering
\caption{Impact of temporal fusion strategies on UCF-Crime and XD-Violence.}
\label{tab:cross_attention_ablation}
\footnotesize % or \scriptsize
\begin{tabular}{lccc}
\toprule
 {Fusion Strategy} &  {UCF AUC (\%)} &  {XD AUC (\%)} &  {XD AP (\%)} \\
\midrule
Concat-Fusion & 86.2 & 88.4 & 76.1 \\
Add-Fusion~\cite{bahdanau2015neural} & 87.0 & 89.1 & 78.2 \\
Cross-Attention (Ours) &  {90.4} &  {92.1} &  {83.7} \\
\bottomrule
\end{tabular}
\end{table}

\paragraph{Memory Bank Size and Diversity.}  
We also change the contextual memory bank size $|\mathcal{M}| \in \{50, 100, 200, 400\}$. Performance increases steadily to $|\mathcal{M}|=200$. then redundancy adds decreasing returns. This is consistent with vector database theory, where diversity beats raw scale.

\begin{table}[h]
\centering
\caption{Influence of memory bank size on UCF-Crime. Saturation after 200 traces.}
\label{tab:memory_size_ablation}{
\begin{tabular}{lcc}
\toprule
 {Memory Size ($|\mathcal{M}|$)} &  {AUC (\%)} &  {F1 (\%)} \\
\midrule
50 & 87.8 & 79.5 \\
100 & 89.2 & 81.4 \\
200 &  {90.4} &  {83.1} \\
400 & 90.3 & 83.0 \\
\bottomrule
\end{tabular}}
\end{table}

\paragraph{Sensitivity to Retrieval Size ($k$).}  
Increasing the number of retrieved traces ($k$) improves robustness to noisy neighbors. Gains plateau beyond $k=5$, reflecting steady contextual retrieval.

\begin{table}[h]
\centering
\caption{Effect of retrieval size ($k$) on UCF-Crime.}
\label{tab:retrieval_k_ablation}{
\begin{tabular}{lcc}
\toprule
 {Top-$k$} &  {AUC (\%)} &  {F1 (\%)} \\
\midrule
1 & 88.1 & 80.2 \\
5 &  {90.4} &  {83.1} \\
10 & 90.2 & 82.9 \\
\bottomrule
\end{tabular}}
\end{table}

\paragraph{Temporal Window Length.}  
We examine temporal receptive field sizes ($W=8,16,32$). Bigger windows enhance contextual reasoning at the cost of increased latency, illustrating the accuracy–responsiveness trade-off.

\begin{table}[h]
\centering
\caption{Impact of temporal window length ($W$) on anomaly detection performance.}
\label{tab:window_ablation}
\begin{tabular}{lcc}
\toprule
\textbf{Window ($W$)} & \textbf{AUC (\%)} & \textbf{F1 (\%)} \\
\midrule
8  & 88.5 & 81.2 \\
16 & 89.6 & 82.4 \\
32 & 90.4 & 83.1 \\
\bottomrule
\end{tabular}
\end{table}

\section{Conclusion and Future Work}
In this paper, we presented TRACE (Temporal Recall with Contextual Embeddings), a new zero-shot video anomaly detection model that integrates motion and appearance through temporal cross-attention across CLIP frame embeddings, complemented by a memory bank of anomalous and non-anomalous traces. We demonstrated that by freezing large pretrained encoders and training light-weight adapter and fusion modules, TRACE maintains CLIP's open-vocabulary alignment while attaining strong performance: high AUC-ROC and F1-score on UCF-Crime and XD-Violence under frame-level annotations, sampling at 30 fps (or the highest available fps per dataset). Our ablation experiments illustrated that cross-attention fusion, size of the memory bank, top-k recall, and non-anomalous trace components significantly impact detection accuracy vs. latency trade-offs. Furthermore, TRACE performs real-time inference rates (on NVIDIA T4) without detection quality compromises, which establishes its operational deployability in surveillance applications.

for future research, investigating long-range temporal modeling (beyond fixed sliding windows) to learn about slow anomaly evolution or anomalies with gradual context drift can be explored, Secondly incorporating auxiliary modalities (e.g., audio, sensor metadata) to strengthen the contextual recall mechanism and minimize false positives in visually ambiguous contexts. Thirdly, probing adaptive or dynamic trace banks: i.e., enabling trace embeddings to be adapted online or relevance-weighted, to accommodate shifting environments (lighting, season, camera view). Fourthly, enhancing threshold calibration and score normalization (e.g., temperature scaling, domain adaptation) such that anomaly scores generalize across datasets without the need for manual tuning can be explored.

\bibliographystyle{unsrtnat}
\bibliography{references}
  %%% Uncomment this line and comment out the ``thebibliography'' section below to use the external .bib file (using bibtex) .


% 	\newblock In {\\em Frontiers in Handwriting Recognition (ICFHR), 2014 14th

% 	\newblock In {\\em Soft Computing and Pattern Recognition (SoCPaR), 2014 6th

% 	\newblock {\\em arXiv preprint arXiv:1804.09028}, 2018.

% \end{thebibliography}

\end{document}